\documentclass[10pt,twocolumn,letterpaper]{article}

\usepackage{cvpr}
\usepackage{times}
\usepackage{epsfig}
\usepackage{graphicx}
\usepackage{amsmath}
\usepackage{amssymb}
\usepackage{textcomp}
\usepackage{relsize}
\usepackage{color}
\usepackage{anyfontsize}
\usepackage{subcaption}
\usepackage[pagebackref=true,breaklinks=true,letterpaper=true,colorlinks,bookmarks=false]{hyperref}

\cvprfinalcopy % *** Uncomment this line for the final submission

\def\cvprPaperID{1} % *** Enter the CVPR Paper ID here

\ifcvprfinal\fi
\begin{document}

%%%%%%%%% TITLE
\title{Technical Report\\
OpenSalicon: An Open Source Implementation of the Salicon Saliency Model}

\author{Christopher Thomas\\
Department of Computer Science\\
University of Pittsburgh\\
{\tt\small chris@cs.pitt.edu}
}

\maketitle
\section{Introduction}
In this technical report, we present our publicly downloadable implementation of the SALICON \cite{huang2015salicon} saliency model. At the time of this writing, SALICON is one of the top performing saliency models on the MIT 300 \cite{mit-saliency-benchmark} fixation prediction dataset which evaluates how well an algorithm is able to predict where humans would look in a given image. Recently, numerous models have achieved state-of-the-art performance on this benchmark, but none of the top 5 performing models (including SALICON) are available for download. To address this issue, we have created a publicly downloadable implementation of the SALICON model described in \cite{huang2015salicon}. It is our hope that our model will engender further research in visual attention modeling by providing a baseline for comparison of other algorithms and a platform for extending this implementation. The model we provide supports both training and testing, enabling researchers to quickly fine-tune the model on their own dataset. We also provide a pre-trained model and code for those users who only need to generate saliency maps for images without training their own model.

\section{Implementation}
\begin{figure*}[t]
\vspace{-4.5em}
\begin{center}
\includegraphics[width=1\textwidth]{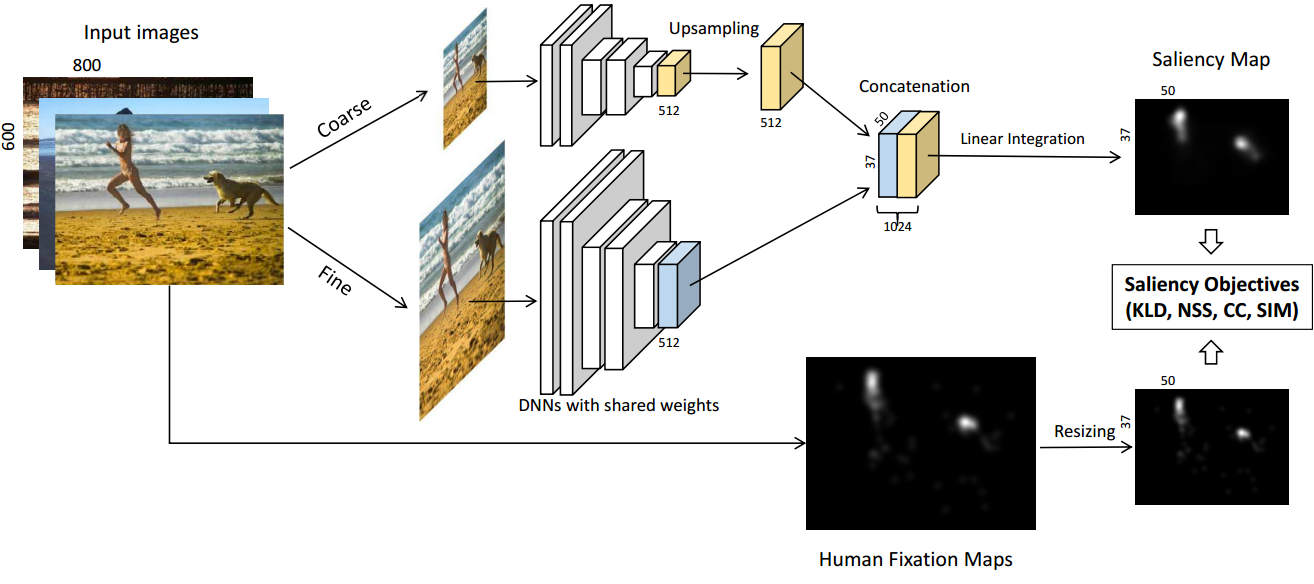}
\end{center}
\vspace{-2em}
\caption{SALICON\cite{huang2015salicon} architecture. The architecture consists of two VGG-16 models which operate on coarse and fine-grained scales of an image. Custom layers are used to provide multiple inputs into the network as well as to perform the upsampling operation for the coarse-grained side of the network.}
\vspace{-1.5em}
\label{salicon}
\end{figure*}
In this section we describe our implementation of the model described in \cite{huang2015salicon}. At a high level, the model consists of two VGG-16 \cite{simonyan2014very} CNNs which run in parallel on the input image at two scales (fine and coarse grain). The model operates in a fully convolutional way in that no ``fully-connected'' layers from the original VGG network are present. Fully-connected layers are replaced by a single ``saliency map'' layer which performs a $1 \times 1$ convolution on the responses from the coarse and fine-grained networks described above (more details are provided below). Our model accepts images of any spatial scale and automatically preprocesses and resizes them to $600 \times 800$ and $1200 \times 1600$ for input into the network. One notable difference between the original implementation described in \cite{huang2015salicon} and our implementation is that the original implementation describes the input to the network as $300 \times 400$ and $600 \times 800$ for the coarse and fine-grained images, respectively. However, using these dimensions does not produce a ``saliency map'' layer of the dimensions described in the paper ($37 \times 50$) for the VGG-16 model and instead produces a result of half that size. In order to produce an output whose dimensions are approximately equal to those described in the paper, our implementation doubles the input dimensions described and produces a saliency map of size $38 \times 50$. The resulting saliency map is then resized to the dimensions of the original input image. A detailed discussion of the design decisions behind this architecture can be found in the original paper.

Our implementation relies on the Caffe deep-learning framework \cite{jia2014caffe}, two custom layers, and several modified network architecture files, along with multiple scripts and functions providing easy utilization for users of the architecture. For ease of implementation and modification, our custom layers are written in Python and accept parameters from Caffe prototxt files. Our model provides three prototxt files. The \texttt{salicon.prototxt} file describes the architecture of our network and is used at test time to produce saliency maps from a given input image. The \texttt{solver.prototxt} file contains parameters and setting used to train our model, such as the learning rate and decay parameters. These parameters are essentially identical to those described in the original paper except for the learning rate which we made slightly smaller to avoid loss divergence during training (which happened on occasion using the original learning rate). To replicate the training procedure described in the original paper, we implemented our own solver in Python (the \texttt{finetune\_salicon.py} file instead of allowing Caffe to automatically solve the network. The final prototxt file we provide is the \texttt{finetune\_salicon.prototxt} file which is used for training the network. This file is a copy of the \texttt{salicon.prototxt} file but also includes a layer for inputting the ground truth fixation map for training purposes as well as a layer for computing the cross entropy loss between the generated saliency map and the ground truth human fixation map for a given image. Additionally, this file provides additional parameters on each of the convolutional layers to control the learning rate and decay rate per layer for both the learned weights and biases. For example, early layers of the CNNs such as the \texttt{conv1\_1} layer have their \texttt{lr\_mult} and \texttt{decay\_mult} parameters set to zero to disable learning. Disabling training of higher layers is standard practice in CNN fine tuning to prevent the network from overfitting. Users of the network wishing to train using different settings can simply change the parameters for each layer to their liking.

Our first custom layer is a utility layer designed for ease of inputting multiple inputs into the network manually. We call this layer the ``CustomData'' layer and it is used to directly feed data into the network from Python without any interference from the Caffe software. Several instances of this layer are used. At test time, we use two instances of the custom data layer. The first feeds fine-scale data into the network and the second feeds coarse data into the network. Users who wish to experiment with different sized inputs can simply change the \texttt{salicon.prototxt} file to declare their desired input data sizes. Those wishing to train the network on different sizes also need to change the dimensions in the \texttt{finetune\_salicon.prototxt} file. An example of this custom layer is below showing the layer is prepared to accept inputs of size $1 \times 3 \times 1200 \times 1600$.
\begin{verbatim}
layer {
  name: "FineCustomData"
  type: "Python"
  top: "fine_scale"
  python_param {
    module: "CustomData"
    layer: "CustomData"
    param_str: "1,3,1200,1600"
  }
}
\end{verbatim}
Our second custom layer is called \texttt{custom\_interpolation\_layer.py} and is used to perform an upsampling. Because the side of the network handling the coarse scale input produces an output half of the size of the output of the fine scale layer, the output of the coarse scale network first needs to be upsampled to the same size as the fine scale output before the two outputs can be combined to avoid dimension mismatch as shown in Fig.\ \ref{salicon}. The capability of resizing blobs in the middle of the network does not exist in Caffe and therefore required a custom layer. Because layers before the custom layer require backpropagation, the custom layer implements both forward and backward capabilities. The forward function of the layer takes an input blob from its bottom and performs a bilinear inpolation using Scipy to resize the bottom blob to the size given in the layer parameter. Users who wish to use different sizes of input data should first change the size of the fine-grained portion of the network and determine what the new size of the \texttt{pool5} layer is. One should then set the parameter of the \texttt{custom\_interpolation\_layer} to that size so that the coarse and fine-grained networks produce output of the same size. Failure of the sizes to agree will cause the software to crash. Using the dimensions described, our network resizes the \texttt{sec\_pool5} blob to size $1 \times 512 \times 38 \times 50$ and sets the result as the layer's ``top'' blob. The backward method of the layer backpropagates the error gradients produced from the loss layer from the ``top'' blob to the ``bottom'' blob. To do this, the layer resizes the ``top'' blob which is of size $1 \times 512 \times 38 \times 50$ to the size of the bottom blob (which is half the size) by performing a bilinear interpolation on the error gradients provided for the ``top'' blob. The declaration and specification of the \texttt{custom\_interpolation\_layer} is shown below.
\begin{verbatim}
layer {
  name: "custom_interpolation_layer"
  type: "Python"
  bottom: "sec_pool5"
  top: "interpolated_data"
  python_param {
    module: "custom_interpolation_layer"
    layer: "custom_interpolation_layer"
    param_str: "1,512,38,50"
  }
}
\end{verbatim}

We use the \texttt{Concat} layer of Caffe to perform concatenation. The result of the process described above produces an output of size $2 \times 512 \times 38 \times 50$. To produce the saliency map, a new convolution layer is added called the \texttt{saliency\_map} layer. The parameters of this layer are set to produce a single channel as output and to use a kernel size of $1 \times 1$. Practically, this consists of learning a weighting for each of the $2 \times 512$ channels and summing over them per pixel. This procedure collapses the output to size $1 \times 1 \times 38 \times 50$ which is the size of the output of the network. The Python program provided then resizes the $38 \times 50$ blob to the size of the original input image using a bilinear interpolation and returns the result to the user.

\section{Using the Pretrained Model}
We trained a model on the OSIE data \cite{xu2014predicting} as in the original SALICON implementation. More details of the training procedure can be found in the training section below. The pretrained model is called \texttt{salicon\_osie.caffemodel}. In order to produce a saliency map for a given image, one first needs to ensure that Caffe and the SALICON folder containing the custom Python modules is on the Python path. This can be done using the \texttt{sys.path.insert(0,path)} command to provide the location of the layers and Caffe installation. The user then creates a ``Salicon'' object, providing the path to the model and prototxt file. This initializes the model and loads the learned weights from the provided file. The user can also set whether they wish to use CPU or GPU computation by editing the Salicon.py file and and using the \texttt{caffe.set\_mode\_cpu()} and \texttt{caffe.set\_mode\_gpu()} commands. Those using GPU computation should also set which GPU device they intend to use by using the command \texttt{caffe.set\_device()} and providing an integer indicating which GPU to use. Once the object is created, saliency maps for images can be produced by using the \texttt{compute\_saliency(path)} command of the \texttt{Salicon} object and providing the absolute path of the image that the user wishes to get the saliency map of. See the below code for an example.
\begin{verbatim}
model = ... # path to salicon_osie.
    caffemodel
prototxtpath = # path to salicon.
    prototxt
salicon = Salicon(
    model=model, 
    prototxtpath=prototxtpath)
image_path = ... # path to image
saliency_map = 
    salicon.compute_saliency(image_path)
# saliency map is the saliency map
    for the image
\end{verbatim}
Users who have trained their own model simply will replace the path to the \texttt{salicon\_osie.caffemodel} (the pretrained model) with their own model. If the network architecture has been changed as well, the path to the modified prototxt file should also be provided. All image preprocessing, such as mean subtraction and image resizing is handled automatically by the \texttt{compute\_saliency} function.

\section{Training}
The training procedure used to train the pretrained model followed the details provided in \cite{huang2015salicon} as closely as possible. As previously described, the learning rate was slightly reduced in this implementation when it was observed that the loss for some models would diverge when the learning rate provided in the paper was used. Users wishing to experiment with different learning parameters should modify the \texttt{solver.prototxt} file. 

Training a model is achieved using the provided \texttt{finetune\_salicon.py} file. The authors in \cite{huang2015salicon} train their model by first initializing their modified network with the weights and biases from the VGG-16 model which is publicly available online. However, because the network architecture of Salicon is not the same as VGG-16, the weights from VGG-16 must be copied to each layer of Salicon one at a time. To avoid users from having to repeat this process, we provide the \texttt{untrained.caffemodel} file. This file contains the Salicon architecture with each layer of the two VGG-16 networks already initialized to the weights and biases of the pretrained VGG-16 network provided by that model's authors. Because the saliency map layer is not in the original VGG-16, its weights are initialized using a Gaussian and its biases are initialized to a constant. Users can use the \texttt{untrained.caffemodel} file to train a model on their own data without using the OSIE data. Those with smaller datasets may also finetune the provided \texttt{salicon\_osie.caffemodel} file instead which has already been trained on the OSIE data.

Training is driven by a custom solver written in Python called \texttt{finetune\_salicon.py}. To train a network on another dataset, ground truth saliency maps (human fixation maps) are required for each image. The data should be organized in two folders, one containing the fixation maps and another containing the images. The fixation maps and the images from which they were produced are required to have the same name. The code reads in each fixation map and then loads in every image. One needs to set the dataset path in the solver's code. The code is parallelized and designed to use a pool of Python workers to load in each fixation map and image and resize it to the appropriate sizes. All images are stored in memory (at their coarse and fine scales) as well as their associated fixation maps, in order to make solving as efficient as possible. Users also need to set the mode of the solver depending on whether they are using a CPU or GPU. This is done the same way as it is for testing (described above). The path to the solver and model file are hardcoded in the code and should be updated depending on whether the user wishes to use different solver parameters or a different model file to initialize the weights from.

Because the original paper does not provide the number of iterations to solve for and only provides an amount of time, the solver runs for 2.5 hours before outputting the trained model and exiting. The provided code assumes that some images will be withheld as a test set, automatically splits the data into train and test set, and only trains on the indexes of the train set, but the code can easily be modified to train on all images. To solve, the solver randomly orders the train images to create a batch. Next, the solver iterates over each image in the batch, one at a time, setting the fine scale, coarse scale, and ground truth and performing backpropagation. It is important to note that the gradients are calculated one image at a time as opposed to the standard minibatch approach of Caffe. This is the way the authors of the original paper train their model because they observed that calculating gradients over minibatches did not improve performance. After all images have been processed, the images are randomly reordered again for the next training epoch. When the timer fires, the solver saves the network to the path provided in the file and exits.

\section{Qualitative Results}
To demonstrate that our pre-trained model produces similar results to the original implementation, we provide qualitative results comparing saliency maps produced by our implementation to those produced by the original authors' implementation. The original authors provide a demo on their website\footnote{\texttt{http://www.salicon.net}} which allows a user to upload an image and obtain a saliency map and provides several example images. We ran our implementation and the demo on several of these example images and provide the results below.
\begin{figure}[t]
\begin{center}
\begin{subfigure}{.33\columnwidth}
\centering
\includegraphics[width=1\columnwidth]{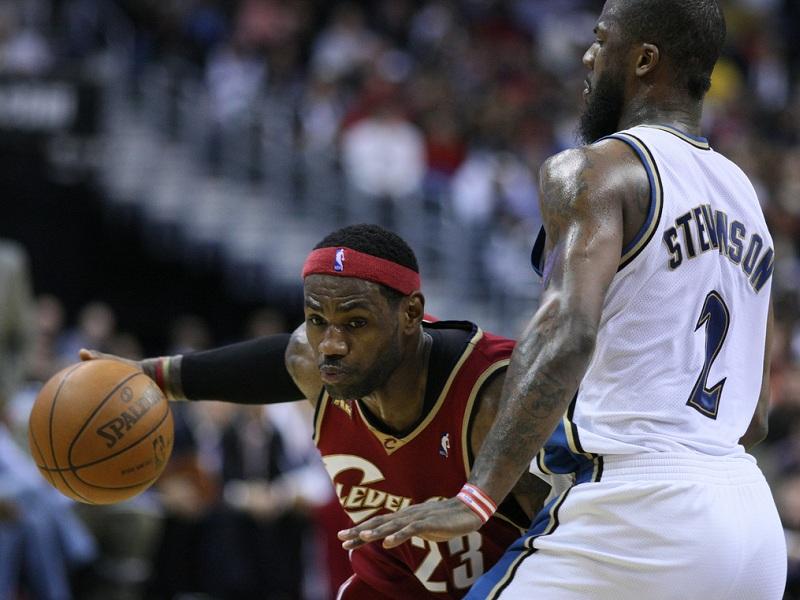}
\caption{Original}
\end{subfigure}%
\begin{subfigure}{.33\columnwidth}
\centering
\includegraphics[width=1\columnwidth]{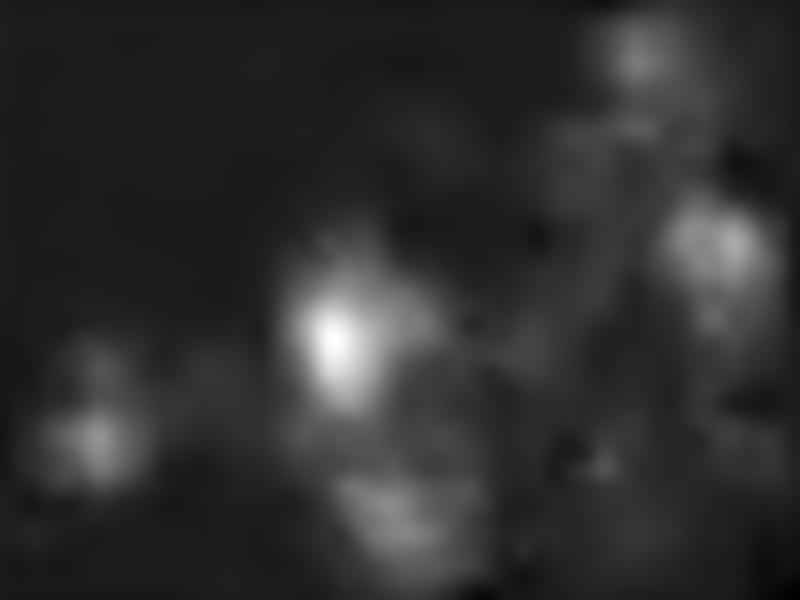}
\caption{Ours}
\end{subfigure}%
\begin{subfigure}{.33\columnwidth}
\centering
\includegraphics[width=1\columnwidth]{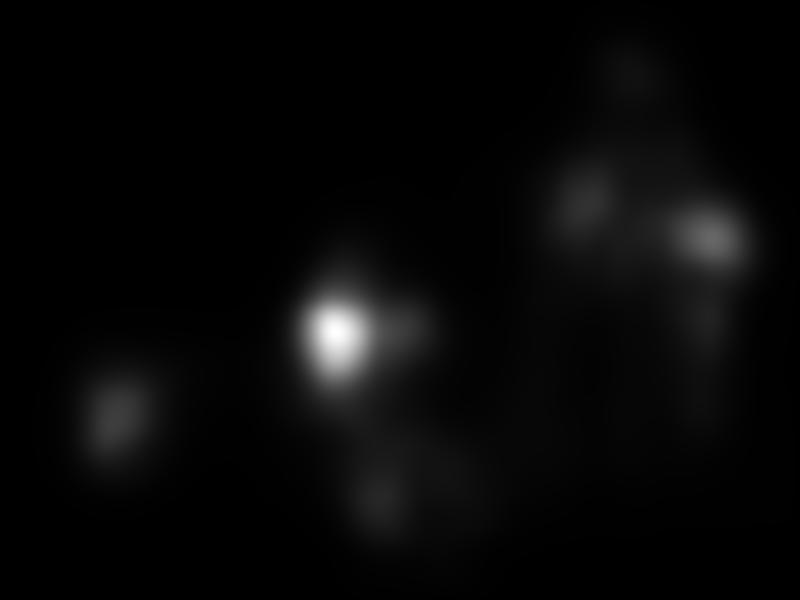}
\caption{Theirs}
\end{subfigure}%
\end{center}

\begin{center}
\begin{subfigure}{.33\columnwidth}
\centering
\includegraphics[width=1\columnwidth]{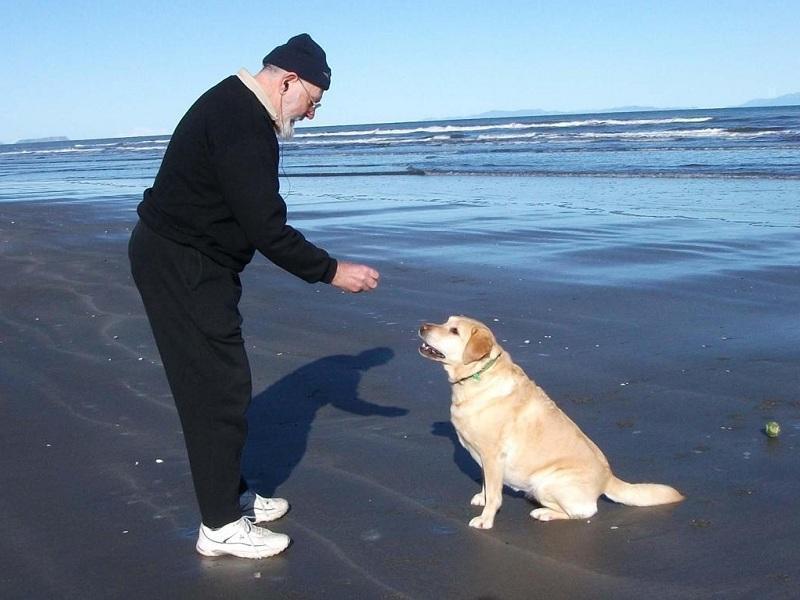}
\caption{Original}
\end{subfigure}%
\begin{subfigure}{.33\columnwidth}
\centering
\includegraphics[width=1\columnwidth]{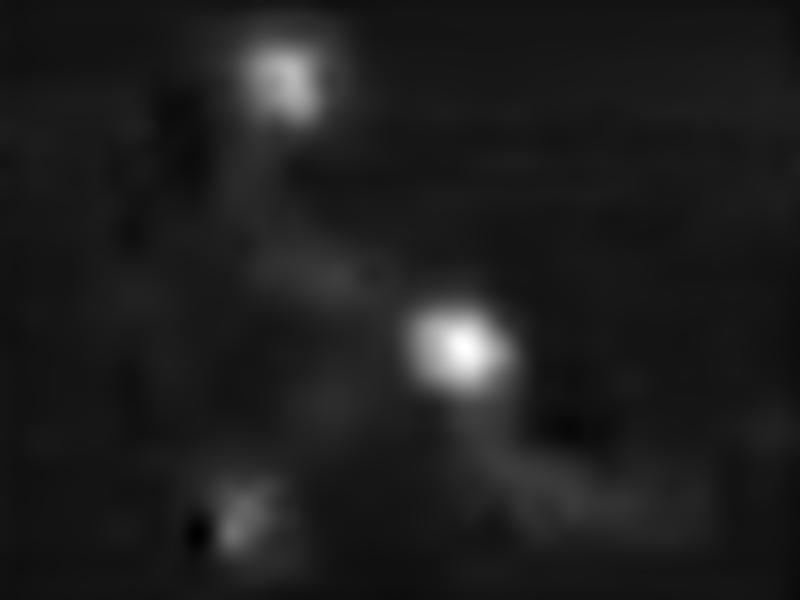}
\caption{Ours}
\end{subfigure}%
\begin{subfigure}{.33\columnwidth}
\centering
\includegraphics[width=1\columnwidth]{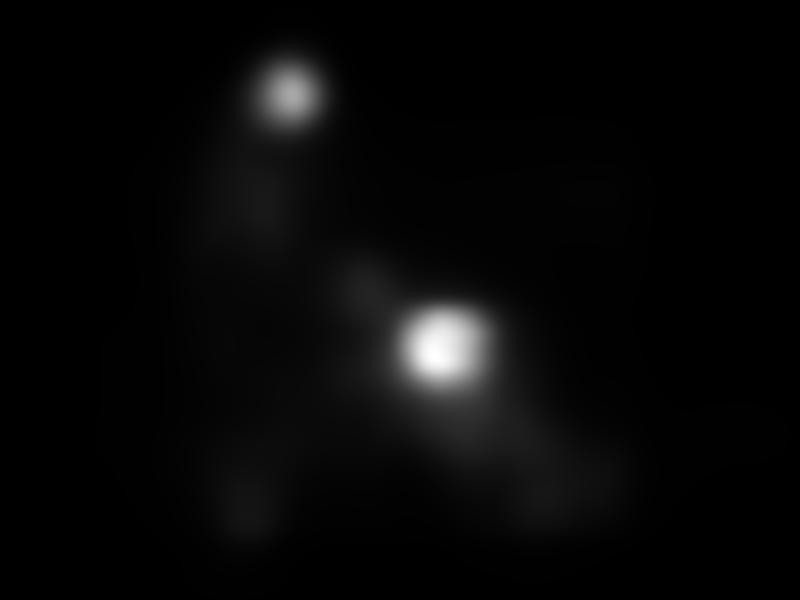}
\caption{Theirs}
\end{subfigure}%
\end{center}

\begin{center}
\begin{subfigure}{.33\columnwidth}
\centering
\includegraphics[width=1\columnwidth]{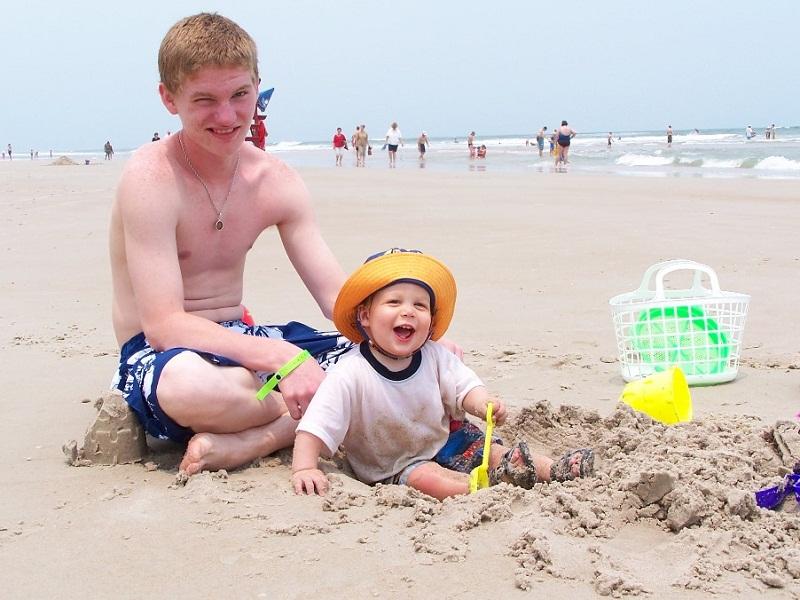}
\caption{Original}
\end{subfigure}%
\begin{subfigure}{.33\columnwidth}
\centering
\includegraphics[width=1\columnwidth]{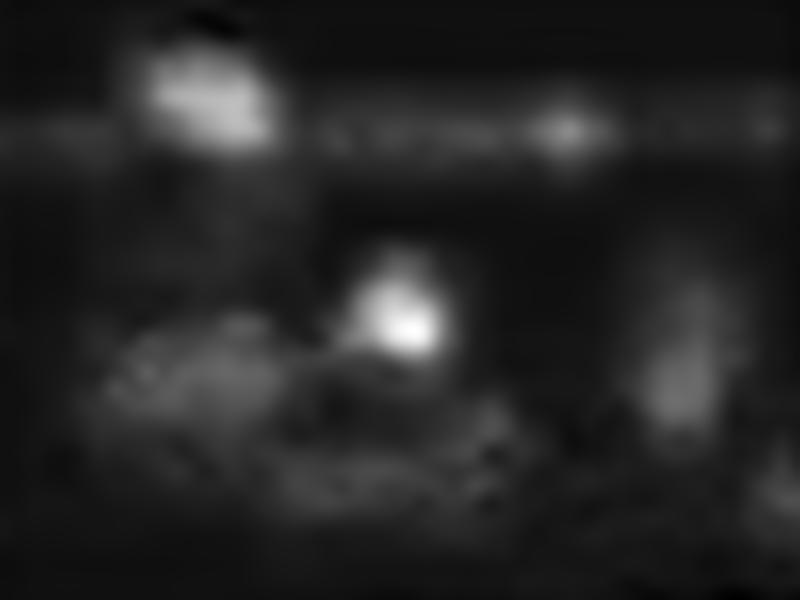}
\caption{Ours}
\end{subfigure}%
\begin{subfigure}{.33\columnwidth}
\centering
\includegraphics[width=1\columnwidth]{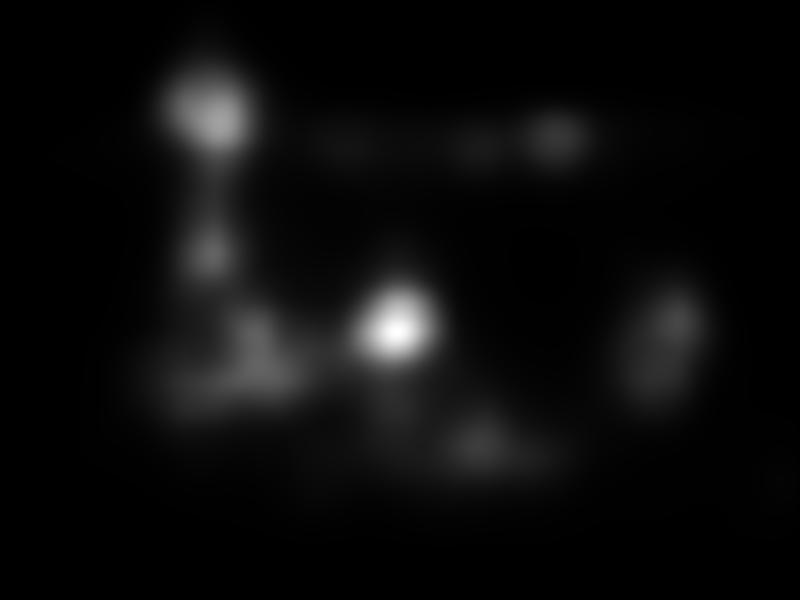}
\caption{Theirs}
\end{subfigure}%
\end{center}

\begin{center}
\begin{subfigure}{.33\columnwidth}
\centering
\includegraphics[width=1\columnwidth]{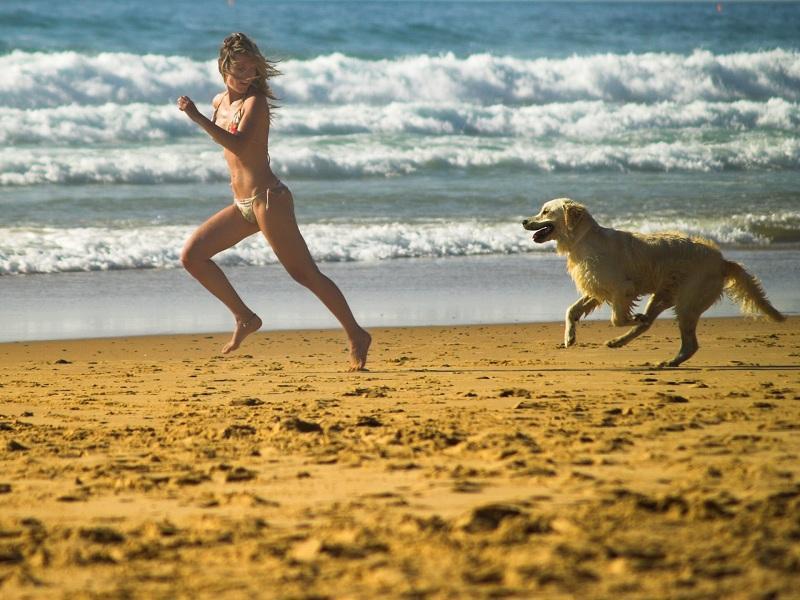}
\caption{Original}
\end{subfigure}%
\begin{subfigure}{.33\columnwidth}
\centering
\includegraphics[width=1\columnwidth]{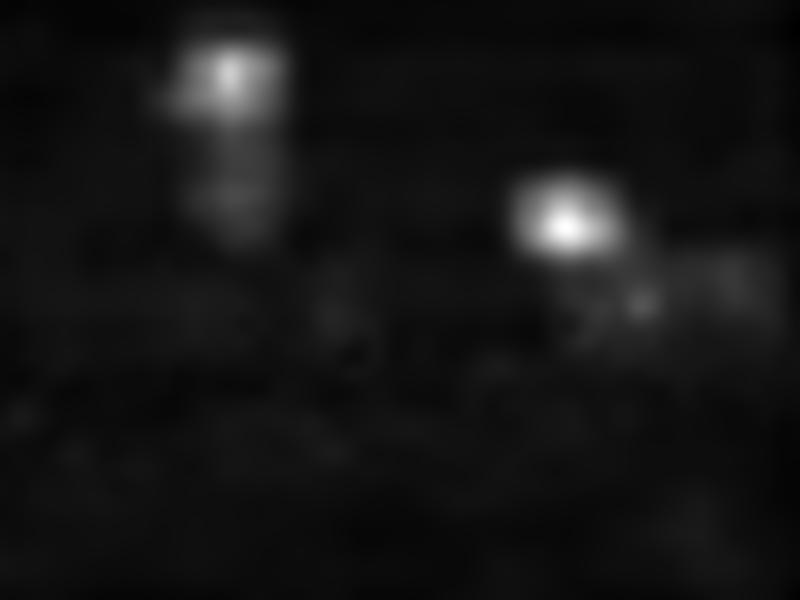}
\caption{Ours}
\end{subfigure}%
\begin{subfigure}{.33\columnwidth}
\centering
\includegraphics[width=1\columnwidth]{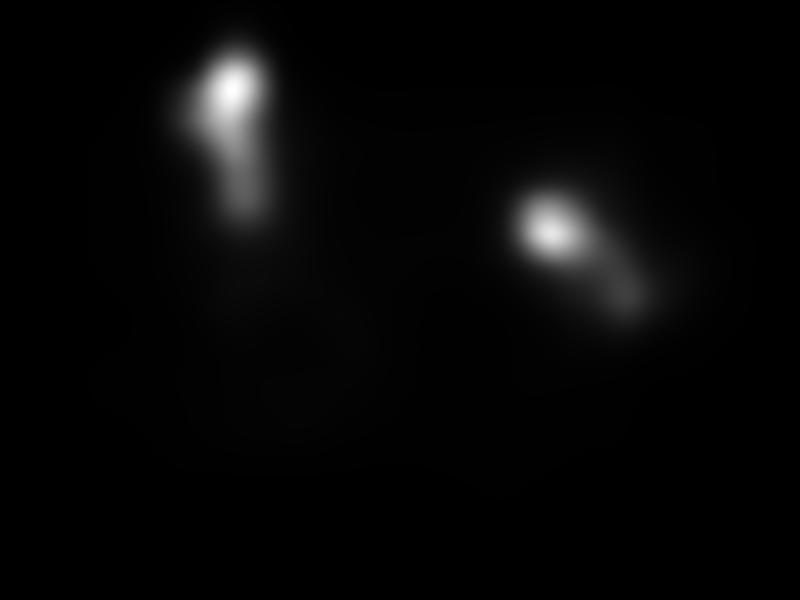}
\caption{Theirs}
\end{subfigure}%
\end{center}
\vspace{-1.5em}
\caption{Qualitative Results. The first column shows the original image. The second column shows the output of our model. The final column shows the results produced by the SALICON demo available online at \texttt{salicon.net}}
\vspace{-1.5em}
\end{figure}

We observe that our results are very similar to those produced by the website. One observation from these results is that the saliency maps produced from the original authors' website appear ``crisper'' and have less background noise. It is our belief that a post-processing step (such as a thresholding) is being performed on the output of the SALICON model to produce tighter, crisper saliency maps. Because this is not described in the paper \cite{huang2015salicon}, we do not perform a final thresholding of the saliency map. Users desiring these crisp saliency maps can perform a thresholding of the saliency map produced by our model and obtain results nearly identical to those produced by the website.

{\footnotesize
\bibliographystyle{ieee}
\bibliography{saliconbib}
}

\end{document}